\documentclass{article}

\usepackage{arxiv}

\usepackage[utf8]{inputenc} 
\usepackage[T1]{fontenc}    
\usepackage{hyperref}       
\usepackage{url}            
\usepackage{booktabs}       
\usepackage{amsfonts}       
\usepackage{nicefrac}       
\usepackage{microtype}      
\usepackage{lipsum}		
\usepackage{graphicx}
\usepackage{doi}
\usepackage[numbers]{natbib}
\usepackage{algorithm}
\usepackage{bm}
\usepackage{booktabs}
\usepackage{bbm}
\usepackage{graphicx}
\usepackage{amsfonts}
\usepackage{amsmath}
\usepackage{algpseudocode}
\usepackage[T1]{fontenc}
\usepackage{amsmath, amssymb}

\title{Contrastive-SDE: Guiding Stochastic Differential Equations with Contrastive Learning for Unpaired Image-to-Image Translation}

\date{} 					

\author{ \href{https://orcid.org/0009-0001-7506-2738}{\includegraphics[scale=0.06]{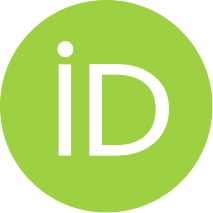}\hspace{1mm}Venkata Narendra Kotyada}\\
	Department of Computer Science \& Engineering\\
    National Institute of Technology,\\
    Andhra Pradesh \\
	\texttt{421202@student.nitandhra.ac.in} \\
	\And
	\href{https://orcid.org/0009-0001-4554-6117}{\includegraphics[scale=0.06]{orcid.pdf}\hspace{1mm}Revanth Eranki} \\
	Department of Computer Science \& Engineering\\
    National Institute of Technology,\\
    Andhra Pradesh \\
	\texttt{422145@student.nitandhra.ac.in} \\
    \And
	\href{https://orcid.org/0000-0003-1743-502X}{\includegraphics[scale=0.06]{orcid.pdf}\hspace{1mm}Nagesh Bhattu Sristy} \thanks{Corresponding author
}\\
	Department of Computer Science \& Engineering\\
    National Institute of Technology,\\
    Andhra Pradesh \\
	\texttt{nageshbhattu@nitandhra.ac.in} \\
}

\renewcommand{\headeright}{Preprint}
\renewcommand{\shorttitle}{Contrastive-SDE}

\begin{document}
\maketitle

\begin{abstract}
Unpaired image-to-image translation involves learning mappings between source domain and target domain in the absence of aligned or corresponding samples. Score based diffusion models have demonstrated state-of-the-art performance in generative tasks. Their ability to approximate complex data distributions through stochastic differential equations (SDEs) enables them to generate high-fidelity and diverse outputs, making them particularly well-suited for unpaired I2I settings. In parallel, contrastive learning provides a powerful framework for learning semantic similarities without the need for explicit supervision or paired data. By pulling together representations of semantically similar samples and pushing apart dissimilar ones, contrastive methods are inherently aligned with the objectives of unpaired translation. Its ability to selectively enforce semantic consistency at the feature level makes contrastive learning particularly effective for guiding generation in unpaired scenarios. In this work, we propose a time-dependent contrastive learning approach where a model is trained with SimCLR by considering an image and its domain invarient feature as a positive pair, enabling the preservation of domain-invariant features and the discarding of domain-specific ones. The learned contrastive model then guides the inference of a pretrained SDE for the I2I translation task. We empirically compare Contrastive-SDE with several baselines across three common unpaired I2I tasks, using four metrics for evaluation. Constrastive-SDE achieves comparable results to the state-of-the-art on several metrics. Furthermore, we observe that our model converges significantly faster and requires no label supervision or classifier training, making it a more efficient alternative for this task. Code is available at \nolinkurl{https://github.com/kvn-257/Contrastive-SDE}
\end{abstract}


\section{Introduction}
Image-to-image translation involves transforming an image from one domain into a corresponding image in another, such as converting a sketch into a realistic photo or a summer scene into a winter one. This task is generally divided into two categories: paired and unpaired settings. In the paired setting, the model learns mapping between source domain and target domain with aligned image pairs, making it a supervised learning problem. The unpaired image-to-image translation setting assumes that source and target domains do not have aligned or matched image pairs, which makes it challenging to learn meaningful mappings without additional constraints or supervision. Early methods for unpaired I2I relied on  GAN based approaches \cite{ref_cyclegan,ref_park2020} for this task. While effective in some cases, GAN-based methods often face limitations such as mode collapse, training instability, and limited output diversity. In the recent years Score-based diffusion models (SBDM)\cite{ref_score_sde} have demonstarted SOTA performance in generating high-quality and diverse images. SBDMs model the data distribution by learning the gradient of the log-density (i.e., the score function) and generate samples by reversing a stochastic diffusion process. This formulation allows for more stable training, avoids mode collapse, and provides greater flexibility in guiding the generation process. To generate images that preserve domain-invariant features like color and texture while discarding domain-specific features like shape, identity, and pose, diffusion models often employ classifier guidance. More recent approaches have proposed energy functions \cite{ref_egsde,ref_sddm} to guide the generation process.
However, an alternative formulation is to treat the objective of preserving domain-invariant features as a representation learning problem, where the model is trained to pull domain-invariant representations together and push domain-specific ones apart. This aligns closely with the principles of contrastive learning, which has proven effective in learning robust and discriminative features without supervision. Motivated by this perspective, we propose a time-dependent contrastive model that learns the domain-invariant features and discard domain-specific features across time steps of the diffusion process. This model is then used to guide the diffusion inference, ensuring that the generated images are aligned with the desired target domain while discarding irrelevant features. A key advantage of this approach is that, by using SimCLR which relies solely on positive pairs and unlabeled data, our contrastive model can be trained efficiently using only domain-invariant features. This eliminates the need for a pretrained or separately trained classifier to extract domain-specific features, thereby reducing training overhead, which simplifies the overall pipeline and leads to faster convergence compared to training a classifier with cross-entropy loss.

We conduct experiments on publicly available datasets including AFHQ and CelebA, on tasks such as Cat → Dog, Wild → Dog, and Male → Female. Our model achieves performance comparable to state-of-the-art methods across several evaluation metrics. To the best of our knowledge, this is the first work to integrate contrastive learning directly into the diffusion process for guiding unpaired image-to-image translation.

\section{Related Works}
Unpaired image-to-image (I2I) translation became widely adopted with methods like CycleGAN \cite{ref_cyclegan}, which introduced cycle-consistency to learn mappings without paired data. It trains two adversarial networks, one mapping $ G:X \rightarrow Y$ and another $F: Y \rightarrow X$, and adds a loss forcing $F(G(x)) \approx x$. Shared latent-space methods like UNIT, on the other hand, align the distributions of two domains into a common embedding. This ensures semantically related images end up with similar encodings. MUNIT goes further by splitting the representation into a domain-invariant content code and a domain-specific style code. This decomposition allows the model to generate multiple diverse outputs from a single input by varying the style vector. Although GAN-based techniques such as MUNIT \cite{ref_munit} and DRIT \cite{ref_drit} enhanced diversity and style transfer, they tended to be unstable and lack expressiveness.

However, most of these methods lack direct control over domain-invariant features to be retained and may, in some cases, transfer unwanted domain-specific artifacts. SDEdit \cite{ref_sdeit} provided a framework for image editing by partially denoising and then re-synthesizing images, while ILVR \cite{ref_ilvr} enabled precise control by conditioning on low-frequency content during the diffusion process. EGSDE \cite{ref_egsde} introduced energy-guided stochastic differential equations, employing an energy function trained on both source and target domains to guide the inference process of a pretrained SDE. SDEdit and ILVR, though powerful, primarily focus on editing or conditional synthesis rather than learning domain-invariant features for unpaired translation. Methods like SDEdit and EGSDE have attempted to bridge gaps by refining the generative process but still fall short in achieving deliberate domain-invariant feature retention. More recently, diffusion models such as Palette \cite{ref_palette} and Score-Decomposed Diffusion Models (SDDM) \cite{ref_sddm} have established state-of-the-art performance on unpaired I2I tasks with improved fidelity and diversity. 

Concurrently, contrastive learning has emerged as a building block for representation learning, and SimCLR \cite{ref_simclr} showed that data augmentation and a contrastive loss can provide robust and generalizable visual features. Supervised contrastive learning \cite{ref_supcon} further used label information to improve feature quality, with applications to multiple downstream tasks. However, most contrastive learning research has focused on representation learning rather than generative modeling, with limited work on its application to image translation or diffusion processes for removing domain-specific bias. Recently, studies have begun exploring the integration of contrastive learning with diffusion models to improve domain-invariant feature preservation and translation quality \cite{ref_clgldm}. While their approach incorporates contrastive loss within a latent diffusion model to guide image-to-image translation, it does not explicitly integrate time-dependent contrastive learning into the diffusion process. This highlights a gap in the literature: the absence of approaches that use time-dependent contrastive learning to supervise generation directly across diffusion timesteps. Addressing this, our work proposes a framework that guides the diffusion process using domain-invariant features learned contrastively over time, without relying on classifier-based supervision or handcrafted energy functions.

\section{Preliminary}

\subsection{Score-based Diffusion Models}
Score-based diffusion models (SBDM) \cite{ref_score_sde}  gradually add noise to data through a forward diffusion process $\{y_t\}_{t\in T}$, and then learn to reverse this process to recover the original data. Let $p(y_0)$ denote the unknown data distribution over $\mathbb{R}^d$ 
  The general form of forward SDE is 
\begin{equation}
dy = f(y, t)\,dt + g(t)\,dw(t)
\end{equation}

where $f: \mathbb{R}^D \rightarrow \mathbb{R}^D$ is the drift coefficient, $g \in \mathbb{R}$ is the diffusion coefficient and $w$ is the Standard Weiner Process. The functions $f(y,t)$ and
$g(t)$ define the noise schedule and govern the perturbation kernel $q_{t|0}(y_t|y_0)$, which describes the forward noising process from time 0 to $t$. A common choice for $f(y,t)$ is an affine function, which leads to a linear Gaussian kernel that enables efficient one-step sampling. 

Let $p_t(y)$ denote the marginal distribution of the forward SDE at time 
$t$, the closed form of reverse SDE is
\begin{equation}
    dy = \left[ f(y, t) - g(t)^2 \nabla_{y} \log p_t(y) \right] dt + g(t)\, d\overline{w}
    \label{reverse}
\end{equation}
where $\overline{w}$ is a reverse-time standard Wiener process, $\nabla_{y} \log p_t(y)$ is the score function that should be approximated using a parameterized score-based model $s_\theta(y,t)$ i.e. $s_\theta(y,t) \approx \nabla_{y} \log p_t(y)$. Thus the Eq. (\ref{reverse}) can be written as
\begin{equation}
    dy = \left[ f(y, t) - g(t)^2 s_\theta(y,t) \right] dt + g(t)\, d\overline{w}
    \label{diffeq}
\end{equation}

Eq. (\ref{diffeq}) can be solved using various SDE solvers to generate images. In \cite{ref_score_sde}, the authors discretize the equation using the Euler-Maruyama method, the solution is given by

\begin{equation}
    y_{t-1} = y_t - [f(y_t, t) - g(t)^2 s_\theta(y_t, t)] + g(t)z, \quad z \sim \mathcal{N}(0, I).
    \label{sol}
\end{equation}

\subsection{Contrastive Learning}

Contrastive learning aims to learn a representation function $f_\theta(\cdot)$ by bringing similar (positive) samples closer and pushing dissimilar (negative) ones apart in the embedding space. Formally, given two views $x_i$ and $x_j$
the contrastive loss is

\begin{equation}
    \mathcal{L}_{\text{contrast}} = -\log \frac{\exp(\text{sim}(f_\theta(x_i), f_\theta(x_j))/\tau)}{\sum_{k=1}^{2N} \mathbbm{1}_{[k \ne i]} \exp(\text{sim}(f_\theta(x_i), f_\theta(x_k))/\tau)}
\end{equation}

where sim$(\cdot,\cdot)$ denotes cosine similarity and $\tau$ is a temperature hyperparameter.
\subsubsection{Unsupervised Contrastive Loss}
SimCLR \cite{ref_simclr} introduces the NT-Xent loss (Normalized Temperature-scaled Cross Entropy Loss), operates on positive pairs \((\tilde{x}_i, \tilde{x}_j)\) of augmented views derived from the same data point. It aims to maximize the similarity between the representations \((z_i, z_j)\), where $z = f_\theta(x)$ is the projection of the encoder output, and considers all other pairs as negatives. The loss is given by

\begin{equation}
    \mathcal{L}_{\text{NT-Xent}}  = - \log \frac{\exp(\text{sim}(z_i, z_j)/\tau)}{\sum_{k=1}^{2N} \mathbbm{1}_{[k \neq i]} \exp(\text{sim}(z_i, z_k)/\tau)}
\end{equation}

\subsubsection{Supervised Contrastive Loss}

SupCon \cite{ref_supcon} modifies the contrastive loss to build upon label information by using all positives in the batch with the same label:
\begin{equation}
    \mathcal{L}_{\text{sup}} = - \sum_{i=1}^N \frac{1}{|P(i)|} \sum_{p \in P(i)} \log \frac{\exp(\text{sim}(z_i, z_p) / \tau)}{\sum_{a \in A(i)} \exp(\text{sim}(z_i, z_a) / \tau)}
\end{equation}
where \(P(i)\) and \(A(i)\) denote positive and all other samples for the anchor \(i\). SupCon is effective in fully supervised settings like fine-grained classification, where class boundaries are subtle and the model must learn differences within the same class.
Given the absence of labeled data for domain-invariant and domain-specific features, we address this problem in an unsupervised setting using the NT-Xent loss.

\section{Method}
We propose a time-dependent contrastive learning based method in which a neural network $F$ is trained to capture domain-invariant features and discards domain-specific features from an image using contrastive NT XTent loss. This network is used to guide the inference of SDE through a guidance function $Q$. 

\subsection{Training the Contrastive Model}
\begin{figure}
\centering
\includegraphics[width=5cm]{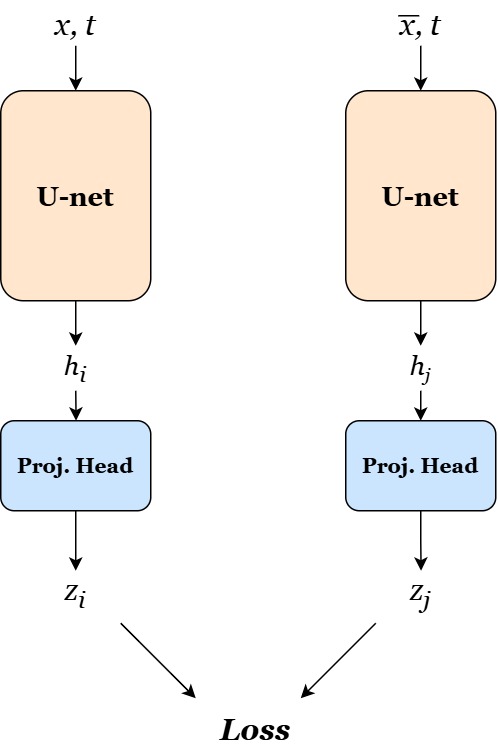}
\caption{Architecure of contrastive model $F$} \label{fig1}
\end{figure}

Intuitively, the ideal setup for a contrastive learning problem is to treat the image and its domain-invariant feature as a positive pair, and the image and its domain-specific feature as a negative pair. However, the NT-Xent loss used in SimCLR has a limitation—it treats all samples in the batch other than the anchor's positive as negatives, without explicitly distinguishing domain-specific from domain-invariant features. In our case, we assume that, when trained on domain-invariant feature, the contrastive model implicitly suppresses domain-specific features by pushing them apart in the embedding space. Initially, we train this U-Net-based architecture $F$ conditioned on the diffusion timestep using a learned time embedding. As depicted in Figure \ref{fig1}, this architecture, comprising a U-Net architecture with residual and attention blocks across multiple resolutions, extracts hidden representations $h_i$ and $h_j$ for the input image $x$ and its low-pass filtered version $\bar{x}$, respectively. Subsequently, a projection head generates contrastive representations $z_i$ and $z_j$. These representations are used to calculate NT XTent loss (\ref{loss}) for training.

\begin{equation}
    \mathcal{L} = \frac{1}{2K} \sum_{k=1}^K \left[ \ell(2k-1, 2k) + \ell(2k, 2k-1) \right] 
    \label{loss}
\end{equation}
 
\begin{equation}
    \ell_{i,j} = - \log \frac{\exp(\text{sim}(z_i, z_j) / \tau)}{\sum_{k=1}^{2K} \mathbbm{1}_{[k \ne i]} \exp(\text{sim}(z_i, z_k) / \tau)}
\end{equation}

where $\text{sim}(u, v) = \frac{u^T v}{\|u\| \|v\|}$ denotes the cosine similarity between two representation vectors $u$ and $v$. $z_i$ and $z_j$ are the learned representation vectors for the positive pair. $\tau$ is a temperature hyperparameter. 2$K$ is the total number of samples in the batch.
$\mathbbm{1}$ is an indicator function that is 1 if $k=i$ and 0 otherwise. Training this contrastive model is relatively easier than training a classifier from scratch for guidance.

\subsection{Guiding Diffusion}
Contrastive-SDE models the conditional distribution $p(y_0|x_0)$ by combining a pre-trained SDE with a guidance function $\mathcal{Q}(y,x,t)$ obtained from network $F$, under mild regularity conditions. 
 The guided reverse time SDE is given by 
\begin{equation}
\mathrm{d}y = \left[ f(y, t) - g(t)^2 \left( s_\theta(y, t) - \nabla_{y} \mathcal{Q}(y, x_0, t) \right) \right] \mathrm{d}t + g(t)\, \mathrm{d}\bar{w},
\label{consde}
\end{equation}
  where $\overline{w}$ is a reverse-time standard Wiener process, $s_\theta(y,t)$ is the parameterized score-based model in the pretrained SDE. $\mathcal{Q}(y, x, t): \mathbb{R}^d \times \mathbb{R}^d \times \mathbb{R} \rightarrow \mathbb{R}^d$ is the guidance function that guides the inference.  
\begin{equation}
\mathcal{Q}(y, x, t) = -\lambda S(y, x, t)
\end{equation}
where $S(y,x,t)$ represents the similarity score function that measures the similarity between source image $y$ and generated sample $x_t$ at time $t$. This score is computed based on the hidden representations extracted by the contrastive model $F$. Specifically, if $h_t$ and $h_0$ are the hidden representations of shapes $C \times M \times N$, obtained from network $F$ for $x_t$ and $y$ respectively.
\begin{equation}
        h_t = F(x_t, t) , \,  h_0 = F(y, t)
\end{equation}
then $S(y,x,t)$ can be expressed as          
\begin{equation}
S(y, x_t, t) = \frac{1}{MN} \sum_{m=1}^{M} \sum_{n=1}^{N} \frac{{h_t^{mn}}^\top h_0^{mn}}{\|h_t^{mn}\|_2 \, \|h_0^{mn}\|_2}
\end{equation}

where $h^{mn}$ denote the channel-wise feature at spatial position $(m, n)$. Alternatively, we can also choose negative squared $L2$ distance as the similarity score function, $S(y,x,t)$ can be expressed as

\begin{equation}
    S(y, x_t, t) = -\left\| h_0 - h_t \right\|_2^2
\end{equation}
We conduct an ablation study to compare the effectiveness of the two similarity functions in guiding the model. Similar to Eq. (\ref{sol}), solution for Eq. (\ref{consde}) becomes
\begin{equation}
    y_{t-1} = y_t - [f(y_t, t) - g(t)^2 (s_\theta(y_t, t) - \nabla_{y} \mathcal{Q}(y_t, x_0, t))] + g(t)z, \, 
\end{equation}
\noindent\[z \sim \mathcal{N}(0, I).\]
\begin{algorithm}
\caption{Contrastive-SDE for unpaired image-to-image translation}
\begin{algorithmic}[1]
\Require Source image $\bm{x}_0$, hyper-parameter $\lambda$, initial time $P$, denoising steps $R$, similarity function $\mathcal{S}(\cdot,\cdot)$, score function $s(\cdot,\cdot)$ 
\State $\bm{y} \sim q_{P|0}(\bm{y}|\bm{x}_0)$ \Comment{Start point}
\State $l \gets \frac{P}{R}$
\For{$i = R$ to $1$}
    \State $t \gets il$
    \State $\bm{x} \sim q_{t|0}(\bm{x}|\bm{x}_0)$ \Comment{Sample perturbed source image}
    \State $\mathcal{Q}(\bm{y}, \bm{x}, t) \gets  - \lambda \mathcal{S}(\bm{y}, \bm{x}, t)$
    \State $\bm{y} \gets \bm{y} - [f(\bm{y}, s) - g(\bm{t})^2 s(\bm{y}, t) - \nabla_{\bm{y}} \mathcal{Q}(\bm{y}, \bm{x}, t)] l$
    \If{$i > 1$}
        \State $\bm{z} \sim \mathcal{N}(\bm{0}, \bm{I})$
    \Else
        \State $\bm{z} \gets 0$
    \EndIf
    \State $\bm{y} \gets \bm{y} + g(t)\sqrt{l}\bm{z}$
\EndFor
\State $\bm{y}_0 \gets \bm{y}$
\State \Return $\bm{y}_0$
\end{algorithmic}
\end{algorithm}

\section{Results \& Discussions}

\subsubsection{Datasets} We evaluate our method using the CelebA-HQ \cite{ref_karras2018} and AFHQ \cite{ref_starganv2} test sets, with all images resized to 256×256. Dataset details are as follows:
(1) CelebA-HQ \cite{ref_karras2018}: Contains 2000 high-quality human face images (1000 male, 1000 female). We perform translation from Male → Female, using male images as the source domain. (2) AFHQ \cite{ref_starganv2}: Comprises 1500 high-resolution animal face images: 500 each for cats, dogs, and wild animals. We conduct translations Cat → Dog and Wild → Dog, using Cat and Wild as source domains respectively.

\subsubsection{Evaluation Metrics} We assess our results based on two factors: realism and faithfulness. Realism of translated images is evaluated using the widely-used Fréchet Inception Distance (FID) \cite{ref_heusel2017} against the target domain, while faithfulness is quantified by the L2 distance, PSNR, and SSIM \cite{ref_wang2004} between input-output pairs.

\subsubsection{Implementation} We train the model F for 2K iterations with a batch size of 32, a learning rate of 3e-4, and a weight decay of 0.05 on AFHQ and CelebA. For guiding all unpaired I2I translation tasks, we used the default settings ($\lambda$ = 500, P = 0.5T), or other setting to improve FID we use ($\lambda$ = 25, P = 0.6T). All experiments are conducted on single NVIDIA RTX A6000 GPU.

\subsection{Results} 
\begin{figure*}[!htbp]
\centering
\includegraphics[width=\linewidth]{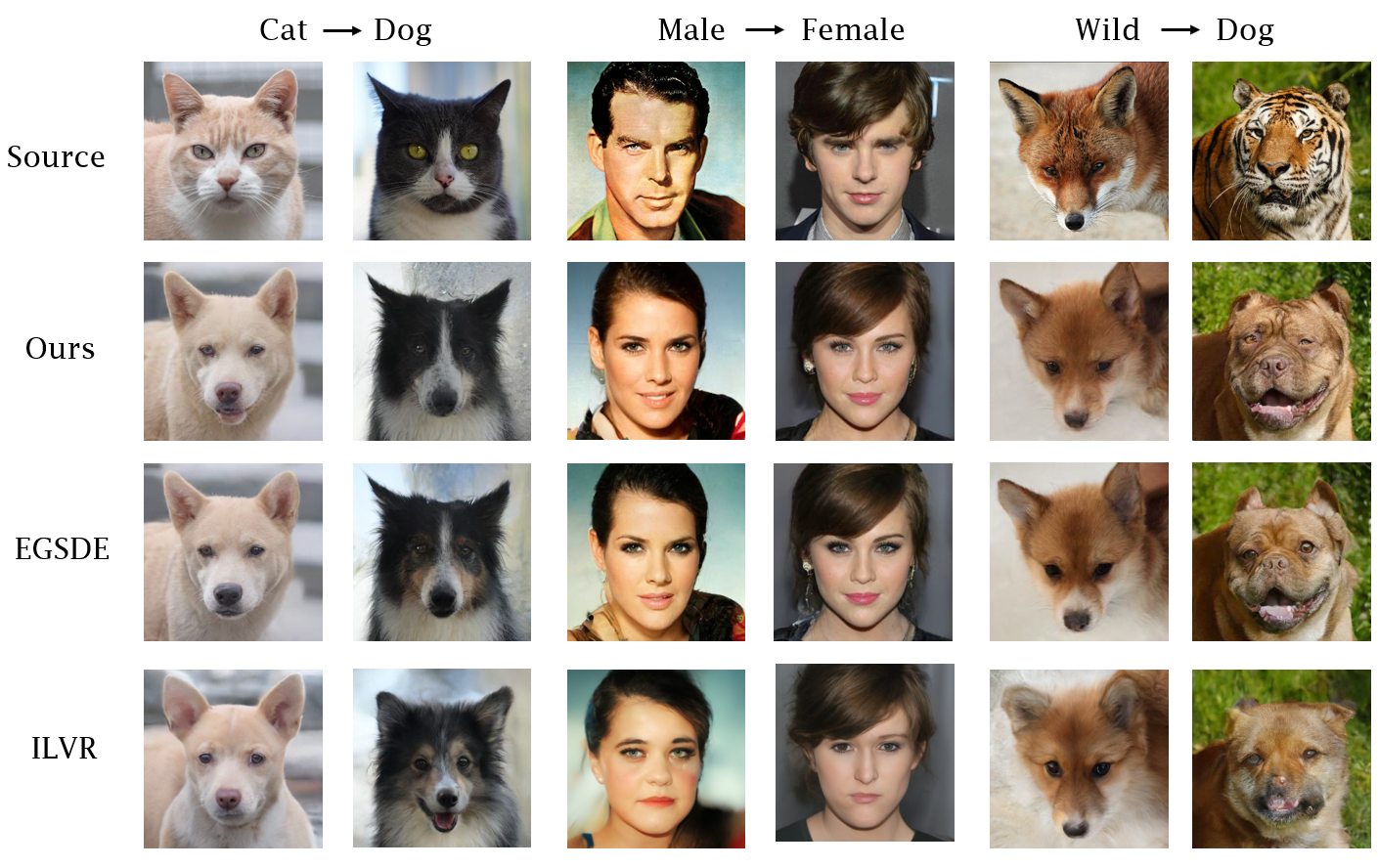}
\caption{Qualitative comparison of Contrastive-SDE with several baselines on three I2I translation tasks.}
\label{fig2}
\end{figure*}

We evaluate Contrastive-SDE against several state-of-the-art (SOTA) baselines: ILVR, EGSDE (with results reproduced from publicly available code), and SDDM (results reported in \cite{ref_sddm}). Quantitative comparison in Table \ref{tab:results} demonstrates that Contrastive-SDE achieves near state-of-the-art performance in faithfulness metrics (L2, PSNR, SSIM), often closely matching EGSDE. For realism, as measured by FID, Contrastive-SDE does not reach the SOTA level but shows superior performance compared to ILVR in default setting. Furthermore, we demonstrate that FID can be improved by adjusting some hyperparameters, offering a flexible trade-off with other metrics. Fig \ref{fig2} presents qualitative results of generated samples across all these baselines. We also compare the training cost of our contrastive model with the classifier-based setup used in EGSDE on the Cat → Dog task. As shown in Table \ref{tab:cost}, EGSDE requires approximately 7 hours of training for 5K iterations, along with labeled data to train the domain-specific classifier. In contrast, our contrastive model converges within 2K iterations in just 2 hours with no labeled data. Additionally, the average training speed is 3.6 seconds per iteration for Contrastive-SDE, compared to 5.06 seconds per iteration for EGSDE. These results show that our method not only removes the need for external classifier supervision but also offers a simpler and more efficient training pipeline.

\setlength{\tabcolsep}{5pt} 
\begin{table*}[h]
\centering
\begin{tabular}{lcccc}
\toprule
Method & FID ↓ & L2 ↓ & PSNR ↑ & SSIM ↑ \\
\midrule
\multicolumn{5}{l}{\textbf{Cat → Dog}} \\
\midrule
ILVR \cite{ref_ilvr} & 74.82 ± 0.88 & 57.04 ± 0.16 & 17.78 ± 0.02 & 0.360 ± 0.001 \\
EGSDE \cite{ref_egsde} & 65.88 ± 0.71 & \textbf{48.68 ± 0.08} & 19.31 ± 0.01 & 0.415 ± 0.001 \\
SDDM* \cite{ref_sddm} & 62.29 ± 0.63 & -- & -- & 0.422 ± 0.001 \\
Contrastive-SDE & 72.61 ± 0.72 & 48.72 ± 0.07 & \textbf{19.31 ± 0.01} & \textbf{0.425 ± 0.001} \\
Contrastive-SDE\textsuperscript{\dag} & \textbf{61.68 ± 0.71} & 59.41 ± 0.10 & 17.54 ± 0.02 & 0.383 ± 0.001 \\
\midrule
\multicolumn{5}{l}{\textbf{Wild → Dog}} \\
\midrule
ILVR \cite{ref_ilvr} & 74.85 ± 0.90 & 63.51 ± 0.10 & 16.84 ± 0.02 & 0.304 ± 0.001 \\
EGSDE \cite{ref_egsde} & 60.13 ± 0.52 & \textbf{55.97 ± 0.08} & \textbf{18.13 ± 0.01} & 0.342 ± 0.001 \\
SDDM* \cite{ref_sddm} & \textbf{57.38 ± 0.53} & -- & -- & 0.328 ± 0.001 \\
Contrastive-SDE & 66.72 ± 0.62  & 56.29 ± 0.11  & 18.08 ± 0.01 & \textbf{0.345 ± 0.001} \\
Contrastive-SDE\textsuperscript{\dag} & 59.86 ± 0.39  & 66.19 ± 0.15 & 16.64 ± 0.02 & 0.304 ± 0.001 \\
\midrule
\multicolumn{5}{l}{\textbf{Male → Female}} \\
\midrule
ILVR \cite{ref_ilvr} &  46.11 ± 0.44  & 52.10 ± 0.09 & 18.60 ± 0.01 & 0.511 ± 0.001 \\
EGSDE \cite{ref_egsde} & \textbf{42.31 ± 0.41} & \textbf{43.00 ± 0.03} & \textbf{20.35 ± 0.01} & 0.574 ± 0.000 \\
SDDM* \cite{ref_sddm} & 44.37 ± 0.23 & -- & -- & 0.526 ± 0.001 \\
Contrastive-SDE & 50.16 ± 0.18 & 44.18 ± 0.04 & 20.11 ± 0.01 & \textbf{0.574 ± 0.001} \\
Contrastive-SDE\textsuperscript{\dag} & 45.15 ± 0.15 & 53.53 ± 0.06 & 18.44 ± 0.02 & 0.533 ± 0.001 \\
\bottomrule
\end{tabular}
\vspace{0.5em}
\caption{Quantitative comparison of Contrastive SDE with several baselines on three I2I translation tasks. The results marked with * are taken from \cite{ref_sddm}. The default setting ($\lambda = 500$, initial time $P = 0.5T$), and $\dag$ denotes the model results with the modified setting ($\lambda = 25$, initial time $P = 0.6T$)}.
\label{tab:results}
\end{table*}

\begin{table*}[h]
\centering
\begin{tabular}{lccccc}
\toprule
Method  &  Training time ↓ & Iterations ↓ & sec/iteration ↓ & batch size \\
\midrule
EGSDE   & 7hr & 5000 & 5.04 & 32  \\
Contrastive SDE  & \textbf{2hr} & \textbf{2000} & \textbf{3.6} & 32  \\
\bottomrule
\end{tabular}
\vspace{0.5em}
\caption{Comparison of training cost for Cat $\rightarrow$ Dog task}.
\label{tab:cost}
\end{table*}

\subsection{Analysis}
The moderate FID performance may be due to using low-pass filtered versions of the images to extract domain-invariant features during training. Low-pass filters may not fully eliminate domain-specific information, which leads to retention in the hidden representations produced by the contrastive model.  We can use more sophisticated domain-invariant feature extractors \cite{ref_higgins2016,ref_kim2018} that could further improve performance. However, based on the results of the faithfulness metrics, we infer that the contrastive model is effectively guiding the diffusion process to preserve domain-invariant features, while also reducing training overhead.The improved per-iteration training time of Contrastive-SDE can be attributed to the use of domain-invariant features during training. Extracting domain-specific features by training a classifier is inherently more challenging than training on domain-invariant features with a contrastive loss, since the latter focuses on learning generalizable patterns that are less complex to model.

\subsection{Ablation}

\begin{figure*}[!htbp]
\centering
\includegraphics[width=13cm]{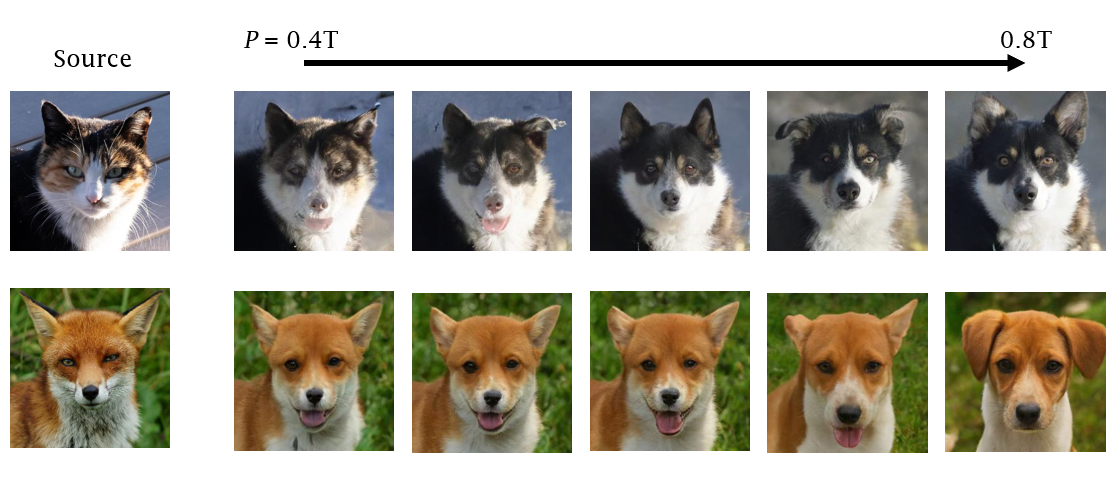}
\caption{Comparison of faithfulness with initial time $P$.}
\label{fig3}
\end{figure*}

\noindent\textbf{Choice of Initial time $\bm{P}$} 
Fig \ref{fig3} illustrates an inverse relationship between $P$ and image faithfulness: as $P$ increases, realism improves, but faithfulness declines.

\noindent\textbf{Choice of Score function $\bm{S}$}
Table \ref{tab:Ablation} shows the effect of score function on metrics on different $\lambda$.

\begin{table}[H]
\centering
\begin{tabular}{lccccc}
\toprule
Simlarity  & $\lambda$ & FID ↓ & L2 ↓ & PSNR ↑ & SSIM ↑ \\
\midrule
Cosine    & 500 &  72.61 & 48.72 & 19.31 & 0.425  \\
Cosine    & 150 &  72.96 & 49.01 & 19.24 & 0.423   \\
NS $L_2$  & 5e-05 & 72.86 & 48.73 & 19.30 & 0.424 \\
NS $L_2$  & 5e-03 & 77.80 & 46.42 & 19.75 & 0.428 \\
\bottomrule
\end{tabular}
\vspace{0.5em}
\caption{Results of both score functions for different $\lambda$ for Cat → Dog I2I task}.
\label{tab:Ablation}
\end{table}

\section{Conclusion}
In this paper, we propose guiding stochastic differential equations with contrastive learning (Contrastive-SDE) for unpaired image to image translation. By training a contrastive model to identify and preserve domain-invariant features, we successfully guided the inference of a pre-trained SDE for effective image translation. Our empirical evaluations across three datasets demonstrate that Contrastive-SDE achieves competitive performance with state-of-the-art baselines and reduces training overhead substantially, with faster convergence than classifier guidance-based models. The limitation of this paper is that we could explore more sophisticated domain-invariant feature extraction techniques or incorporate penalty function in loss for domain-specific features during training, using domain-specific features might increase complexity as it might another model to extract it, but using non-parameterized domain-specific extractors \cite{ref_dalal,ref_ojala} could eliminate the retention of domain-specific features. We will extend our method to other unaligned scenic datasets like Summer2Winter, Horse2Zebra, incorporating the improvements outlined above in future.

\bibliographystyle{unsrtnat}
\bibliography{references}  






\end{document}